\documentclass[10pt,twocolumn,letterpaper]{article}

\usepackage[pagenumbers]{cvpr} 
\usepackage{colortbl} 
\usepackage{multirow} 
\usepackage{booktabs}

\definecolor{cvprblue}{rgb}{0.21,0.49,0.74}
\definecolor{mypink}{rgb}{1.0, 0.08, 0.58}

\usepackage[pagebackref,breaklinks,colorlinks,allcolors=cvprblue]{hyperref}

\title{Volumetrically Consistent 3D Gaussian Rasterization}

\author{Chinmay Talegaonkar, 
Yash Belhe, 
Ravi Ramamoorthi, 
Nicholas Antipa\\
University of California San Diego \\
{\tt\small ctalegaonkar@ucsd.edu}
}

\usepackage{soul}

\definecolor{colorf}{HTML}{FF9396}
\definecolor{colors}{HTML}{FFC991}
\definecolor{colort}{HTML}{FFF6A9}
\newcommand{\boxcolorf}[1]{%
  \begingroup\setlength{\fboxsep}{1pt}%
  \colorbox{colorf}{\hspace*{2pt}\vphantom{Ay}#1\hspace*{2pt}}%
  \endgroup
}
\newcommand{\boxcolors}[1]{%
  \begingroup\setlength{\fboxsep}{1pt}%
  \colorbox{colors}{\hspace*{2pt}\vphantom{Ay}#1\hspace*{2pt}}%
  \endgroup
}
\newcommand{\boxcolort}[1]{%
  \begingroup\setlength{\fboxsep}{1pt}%
  \colorbox{colort}{\hspace*{2pt}\vphantom{Ay}#1\hspace*{2pt}}%
  \endgroup
}
\begin{document}
\maketitle

\begin{abstract}
Recently, 3D Gaussian Splatting (3DGS) has enabled photorealistic view synthesis at high inference speeds.
However, its splatting-based rendering model makes several approximations to the rendering equation, reducing physical accuracy.
We show that the core approximations in splatting are unnecessary, even within a rasterizer;
we instead volumetrically integrate 3D Gaussians directly to compute the transmittance across them analytically.
We use this analytic transmittance to derive more physically-accurate alpha values than 3DGS, which can directly be used within their framework.
The result is a method that more closely follows the volume rendering equation (similar to ray-tracing) while enjoying the speed benefits of rasterization.
Our method represents opaque surfaces with higher accuracy and fewer points than 3DGS.
This enables it to outperform 3DGS for view synthesis (measured in SSIM and LPIPS).
Being volumetrically consistent also enables our method to work out of the box for tomography. We match the state-of-the-art 3DGS-based tomography method with fewer points. Our code is publicly available at:\\ \begingroup
\hypersetup{urlcolor=mypink}
{\small\url{https://github.com/chinmay0301ucsd/Vol3DGS}}
\endgroup

\end{abstract}
\section{Introduction}
Recently, there has been tremendous progress in view synthesis methods using differentiable volume rendering~\cite{mildenhall2020nerf,kerbl20233d,muller2022instant,yu2021plenoctrees,barron2022mipnerf360}.
These techniques can be broadly categorized as rasterization or ray-tracing methods.
Both produce photorealistic results but with distinct trade-offs. 

The speed of rasterization-based methods like 3D Gaussian Splatting (3DGS)~\cite{kerbl20233d} comes at the cost of reduced physical accuracy compared to their ray-tracing-based counterparts.
Rasterization itself makes some unavoidable approximations for faster rendering, like per-tile sorting and no overlap handling (though these can be mitigated at some computational expense~\cite{maule2013hybrid, radl2024stopthepop, hou2024sort}), which 3DGS inherits.
However, 3DGS makes further approximations due to its use of Ellipsoidal Weighted Averaging (EWA) splatting~\cite{zwicker2002ewa}.
These approximations do not merely have a theoretical impact --- they have important practical consequences too.
For example, as we show~\cref{fig:piecewise_constant}, they make it harder for 3D Gaussians to represent opaque surfaces prevalent in 3D scenes.

Recent works have taken inspiration from ray-tracing-based methods to make 3DGS more physically-accurate~\cite{Huang2DGS2024, yu2024gaussian, jiang2024construct}.
However, these methods don't volumetrically integrate 3D Gaussians for color computation. Instead, like 3DGS, they rely on \textit{splatting}, i.e., projecting 3D Gaussians to 2D Gaussians in screen space.

\begin{figure}
    \centering
    \includegraphics[width=\linewidth]{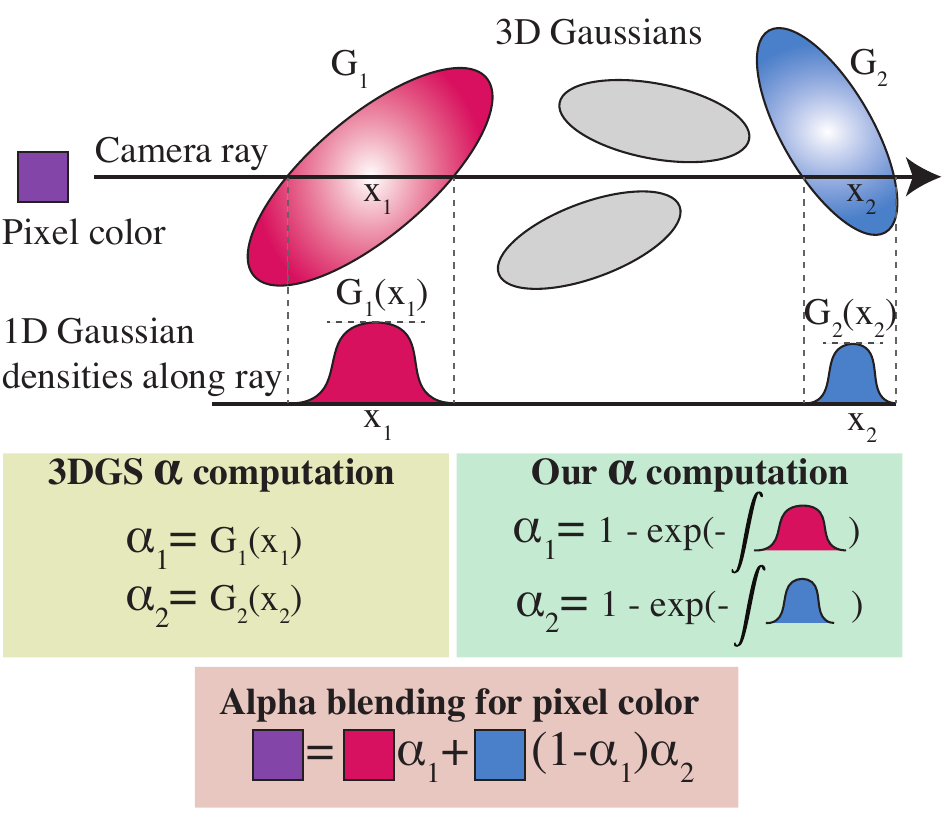}
    \caption{\textbf{Volumetrically consistent $\alpha$ Computation}
    Top row: 3D Gaussians that are not grey contribute to the pixel along the camera ray.
     Next, 1D Gaussian densities along the ray are used to compute $\alpha_i$ values required for color computation.
     3DGS's splatting approximates the volume rendering equation and sets $\alpha_i = G_i(x_i)$, that is the maximum density of the Gaussian long the ray. 
     Our approach instead performs volumetrically consistent $\alpha$ computation by accumulating the density along the ray $\alpha_i = 1 - \exp\left(-
     \int G_i(x)dx\right)$ in accordance with the volume rendering equation, which we derive in~\cref{sec:method}.
     Both methods compute the final pixel color via $\alpha$ blending using their respective $\alpha$ values.
     }
    \label{fig:intuitive_fig}
\end{figure}

We identify that \textit{we do not need splatting}, including all its approximations, to have a fast rasterization-based technique. 
Instead of splatting, we analytically evaluate the volume rendering equation by integrating Gaussians directly in 3D space (assuming correct sorting and no overlap) --- all within a rasterizer. See \cref{fig:intuitive_fig} for a qualitative illustration.  

We first show the volume rendering equation (without splatting approximations) can be expressed as an alpha-blending operation over 3D Gaussians (\cref{subsec:our_alphablend}).
Next, we derive the corresponding alpha values which are more physically-accurate than 3DGS and use them within 3DGS's rasterization framework (\cref{subsec:analytic_transmittance}).
We then show that our alpha values enable representing opaque objects better than 3DGS (\cref{subsec:accurate_alpha_conseq}).

In practice, our method consistently matches or exceeds 3DGS's view synthesis quality (as measured by LPIPS and SSIM) over a wide variety of scenes (\cref{subsec:res_view_synthesis}).

Our method also works out-of-the-box for 3DGS-based tomographic reconstruction and is able to match a recent state-of-the-art method's quality with a lower memory footprint (\cref{subsec:res_tomography}).
\section{Related Work}
\subsection{Volume rendering with ray marching}
Image-based rendering (view synthesis) works have a long history in computer graphics~\cite{gortler1996lumigraph,levoy1996lfr,seitz1996viewmorphing,chen1993viewinterp}.
Recently, NeRF and its followups ~\cite{mildenhall2020nerf, yu2021plenoctrees, barron2021mip, fridovich2022plenoxels, muller2022instant} pioneered using differentiable volume rendering ~\cite{max1995optical} and have achieved photorealistic reconstruction.
These methods use neural networks and grids to represent a scene's volumetric density and color.
Colors along rays are computed evaluating the volume rendering equation using ray marching.
Ray marching typically requires several samples along the ray, each of which involves expensive neural network queries making the (differentiable) rendering process slow.
There has been work on real-time NeRF rendering approaches ~\cite{duckworth2024smerf, reiser2023merf, chen2022mobilenerf, Reiser2024SIGGRAPH, yariv2023bakedsdf, adaptiveshells2023} but these are typically multi-stage approaches, and thus overall still slow. 

\subsection{Rasterizing point-based volumes}
Rasterization-based methods ~\cite{laine2020nvdiffrast,kerbl20233d,zwicker2002ewa} instead represent the scene using explicit primitives, like 3D Gaussians, which are projected onto screen space and composited together.
Tile-based rasterizers like 3DGS are fast, because they exploit the coherence across pixels within the tile by making some assumptions: a) primitives don't overlap; b) sorting order for all primitives is consistent across all pixels in a tile.
3DGS also assumes that blending weights (volumetric transmittance) can be approximated using 2D Gaussians, a result of splatting 3D Gaussians onto the screen~\cite{zwicker2001ewa}.

Very recently, there have been several follow-ups to 3DGS to address these approximations within the rasterization framework.
2DGS~\cite{huang20242d} uses 2D (instead of 3D) Gaussian primitives; they avoid the affine approximation in splatting by computing ray-2D-Gaussian intersection in closed form.
Concurrent work, Gaussian Opacity Fields~\cite{Yu2024GOF} estimates the volume rendering equation in world-space by placing a single sample per Gaussian at its maximum value along a ray.
Others~\cite{jiang24COSV,Zhang2024RaDeGSRD,Dai2024GaussianSurfels,hahlbohm2024htgs} also compute ray-primitive intersections for more accurate depth.

While these methods eliminate the affine approximation, none of them compute physically-accurate transmittance.
Our method, by analytically integrating 3D Gaussians (instead of splatting them onto the screen) is able to do so, resulting in a more physically-based rendering model.
~\citet{hamdi2024ges} replace 3D Gaussians with generalized exponentials which are then splat onto the screen;
their approach enables primitives to be more opaque, resulting in a more compressed representation.
~\citet{radl2024stopthepop} reduce popping artifacts due to inconsistent primitive sorting and ~\citet{hou2024sort} sidestep primitive sorting altogether by using order-independent transparency.
Many recent works~\cite{kheradmand20243d, ye2024absgs, mallick2024taming, kim2024color, bulo2024revising} have improved the densification heuristics in 3DGS and reduced its dependency on initialization.
Our method is a drop-in replacement for the alpha computation in 3DGS, making it compatible with all these recent works as well.
\subsection{Ray-tracing point-based volumes}
Recently, there have been several concurrent works that ray-trace point-based volumes for view synthesis~\cite{moenne20243d, condor2024dontsplatgaussiansvolumetric, mai2024ever}.
These methods vary in their physical-correctness: a) ~\citet{moenne20243d} approximate the volume rendering equation by placing samples at locations of maximum contribution for each primitive along the ray; b) ~\citet{condor2024dontsplatgaussiansvolumetric}, like us, derive a closed-form expression for transmittance across 3D Gaussian and Epanechnikov kernels; c) ~\citet{mai2024ever} compute the volume rendering equation exactly (including sorting and overlap handling) for constant density ellipsoids.
These methods, however, are much slower to train and render than rasterization-based ones, and require specialized hardware for fast ray-tracing.
\subsection{Computed Tomography}
Tomographic reconstruction aims to estimate the underlying 3D density of an object from 2D projections.
Classical analytic methods like FDK~\cite{feldkamp1984practical} invert the Radon transform~\cite{radon1986determination} and are fast but struggle with sparse measurements.

Recently, NeRF-based approaches~\cite{zha2022naf,zang2020tomofluid,zang2021intratomo,cai2024radiative, shen2022nerp} have shown promising results, but are slow due to ray marching.
Several 3DGS-based tomography approaches have been proposed ~\cite{cai2023structure, gao2024ddgs, zha2024r}.
~\citet{zha2024r} are the state-of-the art and achieve high quality by fixing integration bias in 3DGS's Splatting routine. 
\section{Background and Motivation}
In this section we review the volume rendering equation in~\cref{subsec:volume_rendering} and describe all the approximations 3DGS makes to estimate it in~\cref{subsec:3dgs_approx}.
\label{sec:background}
\subsection{Volume rendering}
\label{subsec:volume_rendering}
NeRF-based methods represent a 3D scene as an emissive volume with density (attenuation coefficient) $\sigma(\boldsymbol{x})$ and view-dependent color $c(\boldsymbol{x}, \boldsymbol{d})$. 
The resulting color along a ray $\boldsymbol{r}(t) = \boldsymbol{o} + t\boldsymbol{d}$ (origin $\boldsymbol{o}$ and direction $\boldsymbol{d}$) is given by the volume rendering equation
\begin{align}
C(\boldsymbol{r}) = \int_{0}^{\infty} T(0,t)\sigma(\boldsymbol{r}(t))c(\boldsymbol{r}(t),\boldsymbol{d})dt,
\label{eq:vre}
\end{align}
where $T(a,b)$ is the (exponential) transmittance across the ray segment beginning at $\boldsymbol{r}(a)$ and ending at $\boldsymbol{r}(b)$
\begin{align}
\label{eq:trans_general}
T(a,b) = \exp\left(-\int_{a}^{b} \sigma(\boldsymbol{r}(s))ds\right).
\end{align}
\subsection{Volume rendering of 3D Gaussians}
\label{subsec:3dgs_approx}
We briefly recap the Elliptical Weighted Average (EWA) splatting algorithm~\cite{zwicker2002ewa} used by 3DGS~\cite{kerbl20233d}, and point out the different approximations it makes by \underline{underlining} them.
The density field
\begin{align}
    \sigma(\boldsymbol{x}) = \sum_{i=1}^N \kappa_i G_i(\boldsymbol{x}) 
    \label{eq:gauss_mixture}
\end{align}
is a weighted ($\kappa_i$) sum of unnormalized 3D Gaussians 
\begin{align}
    G_i(\boldsymbol{x}) = \exp\left\{-\frac{1}{2}(\boldsymbol{x} - \boldsymbol{\mu_i})^{T}\boldsymbol{\Sigma_i}^{-1}(\boldsymbol{x}-\boldsymbol{\mu_i})\right\}
    \label{eq:3dgs_pdf}
\end{align}
which are assumed to \underline{not overlap}, have view-dependent color $c_i$, and be \underline{sorted} front-to-back~\cite{zwicker2002ewa}. We also adopt these two assumptions.
To compute pixel color, both EWA and 3DGS, \underline{linearize the exponential} transmittance ($e^{-x} \approx 1-x$), an assumption not strictly necessary for splatting. Our method relaxes this assumption and subsequent assumptions required for splatting. For splatting, both EWA and 3DGS assume Gaussians are \underline{not self-occluding}, resulting in the alpha blending equation describing the color at pixel $p$
\begin{align}
    C(p) = \sum_{i=1}^N c_i\alpha_i\prod^{i-1}_{j=1} (1-\alpha_j).
\label{eq:3dgs_alpha_blend}
\end{align}
Splatting $i$-th 3D Gaussian onto the image plane results in 2D opacity
\begin{align}
    \alpha_i = o_i  \hat{G}_i(p){,}
    \label{eq:3dgs_alpha}
\end{align}
where $o_i$ is the peak opacity, restricted to be in $[0,1]$, of 2D Gaussian $\hat{G}_i(p)$. 
The 2D mean $\mu'_i$ is the projection of 3D mean $\boldsymbol{\mu}_i$ onto the image plane and the 2D covariance is given by
\begin{align}
    \Sigma' = \boldsymbol{JW}\boldsymbol{\Sigma}\boldsymbol{W^{T}J^{T}}. \label{eq:affine_approx}
\end{align}
$\boldsymbol{J}$ is the \underline{affine approximation} (Jacobian) to the camera projection matrix.
For more details please see~\citet{zwicker2002ewa}.
We stress that the values $\alpha_i$ are approximate, even when the sorting and non-overlapping assumptions are met, due to the three additional approximations made by splatting.
We do not perform splatting, thereby relaxing the extra approximations 3DGS entails. 
Instead we compute $\alpha_i$ by analytically evaluating \cref{eq:trans_general}, which accumulates the density $\sigma(\boldsymbol{x})$ along ray $\boldsymbol{r}(t)$.
This results in an efficient rendering method that matches or exceeds state-of-the-art results in both view synthesis and sparse-view tomography. 

\section{Method}
\label{sec:method}
Our method analytically computes the volume rendering equation along a ray for a mixture of 3D Gaussians.
We first describe how this analytic integral (without splatting approximations) can be expressed as an alpha-blending operation in~\cref{subsec:our_alphablend}.
We derive the corresponding alpha values in~\cref{subsec:analytic_transmittance}.
Next, by swapping 3DGS's alpha computation with ours, we show how our method can produce more accurate renderings of opaque objects in~\cref{subsec:accurate_alpha_conseq}.

\subsection{Alpha blending without splatting}
\label{subsec:our_alphablend}
In this section we describe how our method, which analytically integrates transmittance, can also be written as alpha blending due to the properties of exponential transport.
Substituting \cref{eq:gauss_mixture} into \cref{eq:vre} results in the color for a pixel $p$ (with the ray $\boldsymbol{r}(t)$ passing through its center) being
\begin{align}
C(p) &= \sum_{i=1}^N \int_{t_{in}}^{t_{if}} T(0,t)c_i\kappa_iG_i(\boldsymbol{r}(t))dt.
\end{align}
Here $[t_{in},t_{if}]$ are the limits of Gaussian $i$ along the ray.
Assuming the Gaussians are sorted (front to back) and non-overlapping, we can separate out the transmittance due to the previous Gaussians
\begin{align}
C(p) &= \sum_{i=1}^{N}  \left(\prod_{j=1}^{i-1} \overline{T}_j\right) c_i\int_{t_{in}}^{t_{if}} T(t_{in},t)\kappa_iG_i(\boldsymbol{r}(t))dt.
\label{eq:color_temp}
\end{align}
The innermost integral weights the color contribution from Gaussian $i$, accounting for self-occlusion.
The accumulated transmittance across previous Gaussians $j<i$ is
\begin{align}
\overline{T}_j = T(t_{jn}, t_{jf}) = \exp\left(-\int_{t_{jn}}^{t_{jf}}\kappa_jG_j(\boldsymbol{r}(s))ds\right).
\label{eq:trans}
\end{align}
The second equality holds because we assume the Gaussians are non-overlapping.
Note that from the definition of exponential transmittance in~\cref{eq:trans_general} and differentiating, $dT(t_{in}, t) = - T(t_{in}, t) \sigma(\boldsymbol{r}(t))$, which corresponds directly to the integrand in~\cref{eq:color_temp}.  Therefore, this integral results in the $i$-th Gaussian's accumulated opacity, $\alpha_i = 1 - \overline{T}_i$:
\begin{align}
\int_{t_{in}}^{t_{if}} T(t_{in},t)\kappa_iG_i(\boldsymbol{r}(t))dt &= -\int_{t_{in}}^{t_{if}} dT(t_{in},t) \\
&= T(t_{in},t_{in}) - T(t_{in},t_{if}) \\
&= 1 - \overline{T}_i = \alpha_i.
\end{align}
Substituting this result into \cref{eq:color_temp} yields the final alpha-blending equation
\begin{align}
C(p) &= \sum_{i=1}^{N}  c_i\alpha_i \prod_{j=1}^{i-1} (1 - \alpha_j) .
\label{eq:our_alpha_blend}
\end{align}
We observe that this equation is identical to the alpha-blending equation, \cref{eq:3dgs_alpha_blend}, adopted by 3DGS. 
The difference is that \cref{eq:our_alpha_blend} requires accurately evaluating $\alpha_i$, the accumulated density of the $i$-th primitive. 
Note that thus far, we do not rely on any specific property of 3D Gaussians; we only require the primitives $G_i$ to be non-overlapping.
This opens up a design space for other compact kernels which we leave for future work; we use 3D Gaussians.
In the next section, we show that the $\alpha$ values in \cref{eq:our_alpha_blend} can be computed analytically for 3D Gaussians.
The result closely matches ray tracing results (assuming no overlap and correct sorting), but uses rasterization for efficient rendering, see Supp.A.
\subsection{Analytic transmittance computation}
\label{subsec:analytic_transmittance}
The final step required to compute the color $C(p)$ is to estimate the transmittance~\cref{eq:trans}, which requires integrating a 3D Gaussian along a 1D ray.
The corresponding (scaled) 1D Gaussian $g(t)$ in ray-coordinates (i.e. parameterized by the distance $t$, ray origin $\boldsymbol{o}$ and direction $\boldsymbol{d}$) is
\begin{align}
    g_j(t) &= G_j(\gamma_j \boldsymbol{d}) \exp\left\{\frac{-(t - \gamma_j)^{2}}{2\beta_j^{2}}\right\},\label{eq:1d_gauss}
\end{align}
where $G_j(\gamma_j\boldsymbol{d})$ is the maximum value of the Gaussian along the ray, and the 1D Gaussian has mean $\gamma_j$ and variance $\beta_j$  (proof in Supp.B)
\begin{align}
    \gamma_j &= \frac{(\boldsymbol{\mu_j} - \boldsymbol{o})^{T} \Sigma_j^{-1} \boldsymbol{d}}{\boldsymbol{d}^{T}\Sigma_j^{-1}\boldsymbol{d}}, \;\;\;\;\;
    \beta_j = \frac{1}{\sqrt{\boldsymbol{d}^T \Sigma_j^{-1} \boldsymbol{d}}}.
\end{align}
Now, the final integral necessary to compute~\cref{eq:trans} is
\begin{align}
    \int_{t_{jn}}^{t_{jf}}  g_j(t)dt &= \int_{t_{jn}}^{t_{jf}}  G_j(\gamma_j \boldsymbol{d}) \exp\left\{\frac{-(t - \gamma_j)^{2}}{2\beta_j^{2}}\right\}dt, \\
    &= G_j(\gamma_j \boldsymbol{d}) \sqrt{\frac{\pi}{2}}\beta_j\text{erf}\left(\frac{t-\gamma_j}{2\beta_j}\right)\biggr\rvert_{t_{jn}}^{t_{jf}}.
\end{align}
Assuming infinite support, the transmittance is given by
\begin{align}
    \boxed{\;\;\;\overline{T}_j = \exp\left(-\kappa_j G_j(\gamma_j \boldsymbol{d}) \sqrt{2\pi}\beta_j\right).\;\;\;}
    \label{eq:our_trans_final}
\end{align}
Concurrent works~\cite{Yu2024GOF,moenne20243d} also estimate the volume rendering equation by sampling the maximum value along the ray (instead of analytic integration), which approximates transmittance; ~\citet{condor2024dontsplatgaussiansvolumetric} derive the same transmittance expression as us and use it for ray-tracing 3D Gaussians.
\subsection{Benefits of accurate alpha computation}
\begin{figure}[t]
    \centering
    \includegraphics[width=\linewidth]{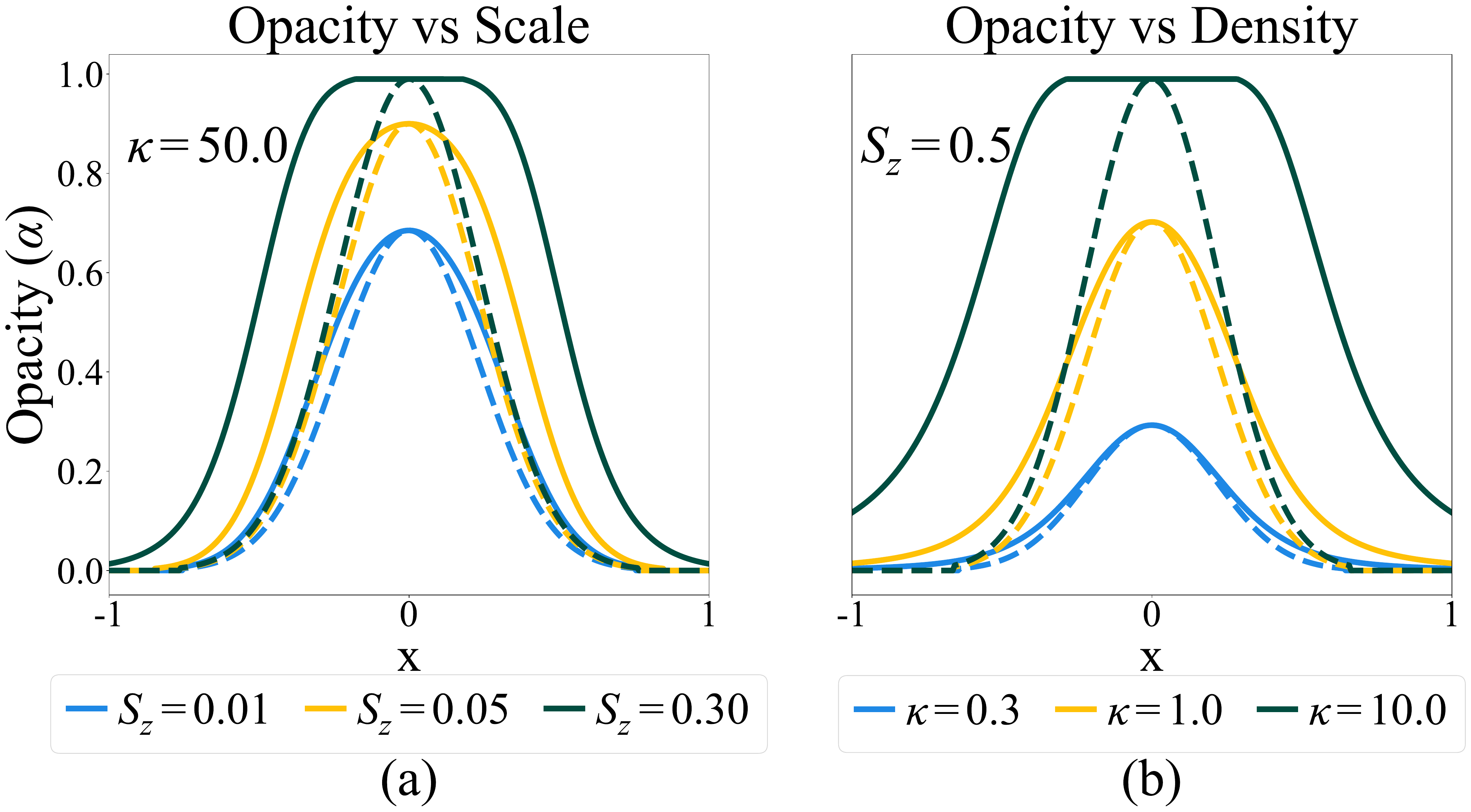}
    \includegraphics[width=\linewidth]{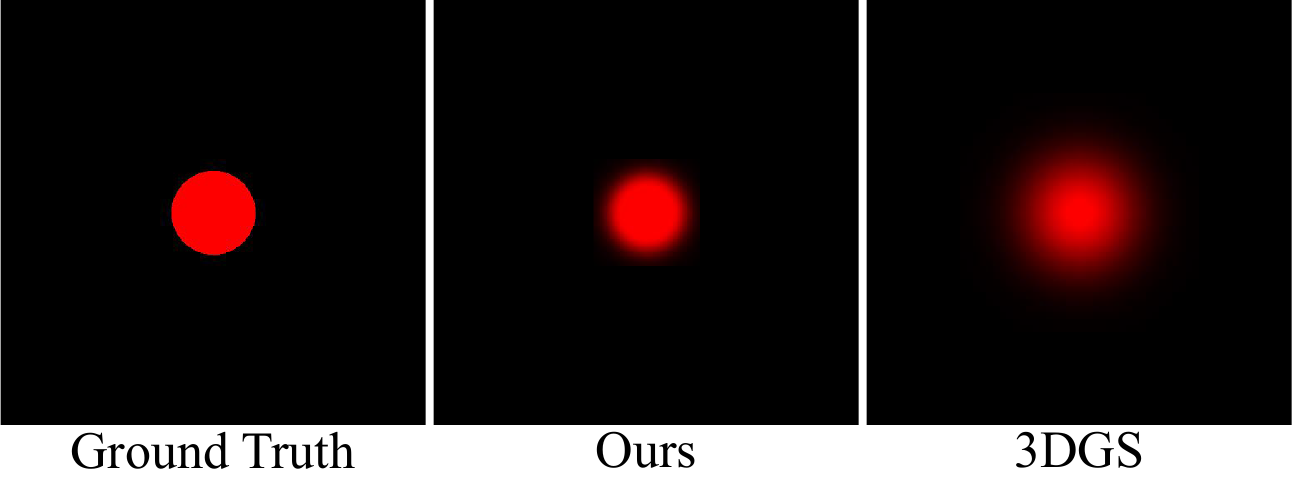}
    \caption{
    \textbf{Opacity as a function of density and scale.}
    \textit{Top row:} 1D cross sections of renderings for our method (solid) and 3DGS (dotted).
    Our method represents opaque objects better by two mechanisms: 1) increasing the scale along the camera ray (a); 2) increasing the volume density (b); both increase the flat region where $\alpha=1$.
    Since 3DGS splats 3D Gaussians onto the image plane as 2D Gaussians, irrespective of the scale (a) or the opacity (b), the cross-sections are all Gaussian with $\alpha=1$ possible only at the center. 
    \textit{Bottom row:} We optimize the parameters of a single 3D Gaussian to fit a circle using 3DGS and our method.
    As we saw in the top row (a) and (b), regardless of scale or opacity, 3DGS can only render a 2D Gaussian in image-space, resulting in a blurry fit of the opaque object. 
    On the other hand, our method adjusts density and scale to produce a more opaque rendering.}
    \label{fig:1d_scale_density}
\end{figure}

\begin{figure}
    \centering
    \includegraphics[width=\linewidth]{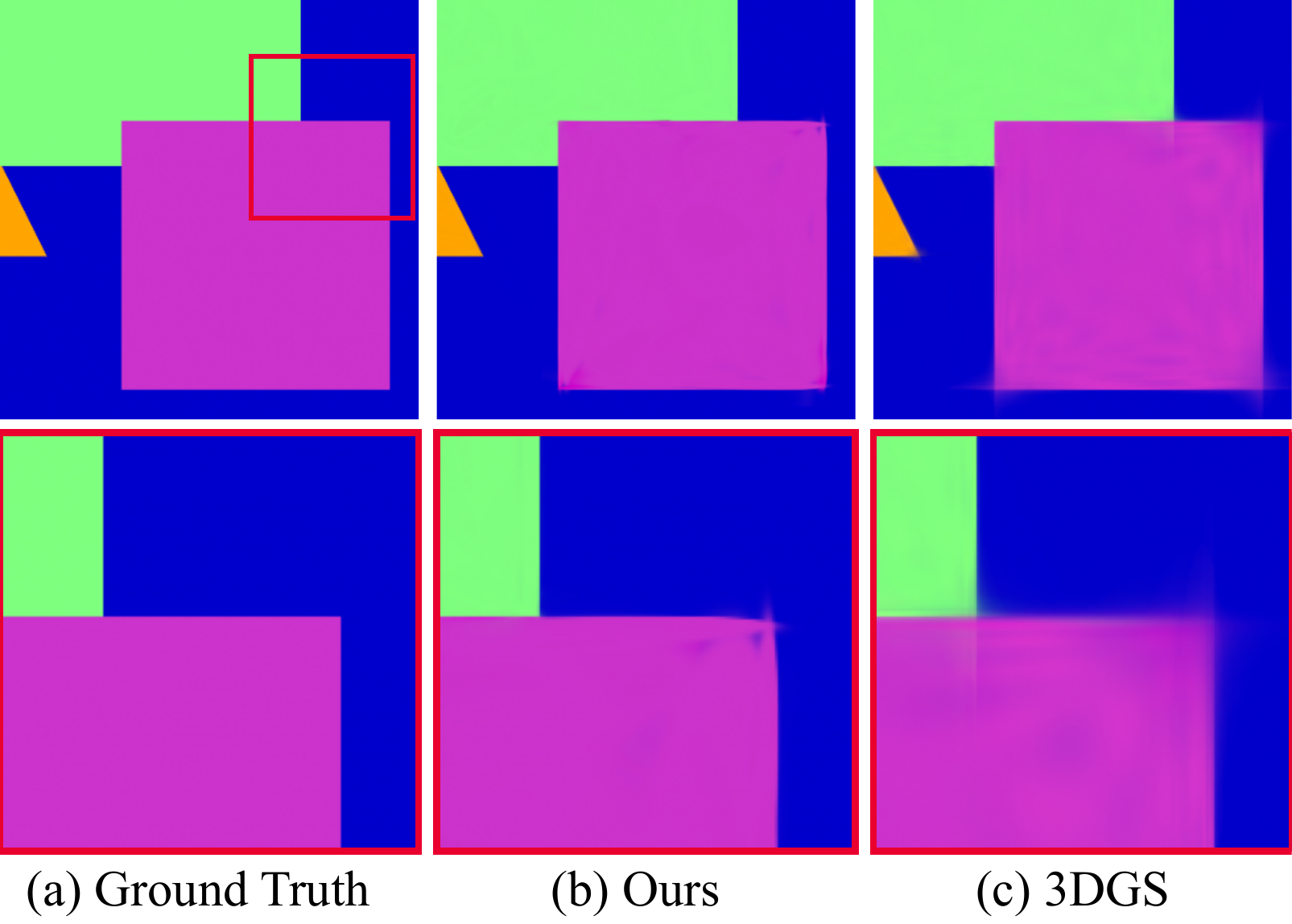}
    \caption{\textbf{Fitting piecewise constant shapes with equal numbers of 3D Gaussians.}
    (Please zoom in.) Our method, by increasing the volume density $\kappa$ can make rendered 3D Gaussians close to opaque (b), which is a closer match to the ground truth (a), with LPIPS $0.005$ (lower is better).
    3DGS splats 3D Gaussians onto the screen as 2D Gaussians, which are only fully opaque at their center; this leads to artifacts especially near edges (see inset in second row) and much worse LPIPS $0.027$.
 }
    \label{fig:piecewise_constant}
\end{figure}

\label{subsec:accurate_alpha_conseq}
Comparing our alpha derivation to 3DGS's, we note that 3DGS makes three extra approximations (see~\cref{subsec:3dgs_approx}): it assumes a) linearization of the exponential; b) no-self occlusion; c) linearization of the covariance.
Each of these make renderings less physically-based and hurt view synthesis quality.

The result of the first two approximations is that for 3DGS $\alpha_i$ is given by a 2D Gaussian, see~\cref{eq:3dgs_alpha}.
This 2D Gaussian only reaches its maximum opacity of 1 at its center (\cref{fig:1d_scale_density} (b), dotted), limiting 3DGS's ability to represent opaque surfaces ($\alpha=1$). 
As a consequence, our method is better at fitting piecewise constant textures than 3DGS for a fixed budget of Gaussians. 
We demonstrate this through examples in \cref{fig:1d_scale_density} and \cref{fig:piecewise_constant}. 
In \cref{fig:1d_scale_density} we plot the opacity, $\alpha_i$, of a volumetrically rendered Gaussian with varying density and scale. 
We observe that when increasing the density $\kappa_i$, our method results in opacity close to 1 over a much larger region (\cref{fig:1d_scale_density} (b), solid), thereby enabling it to better represent opaque surfaces (\cref{fig:1d_scale_density}, bottom row). 
Another consequence of splatting a 3D primitive to 2D is that it discards the z-scale (along the camera ray) of the Gaussian, which makes renderings invariant to it (\cref{fig:1d_scale_density} (a), dotted).
This is physically inaccurate; larger z-scales should increase a Gaussian's contribution, making it more opaque. 
Our method captures this effect, as shown in \cref{fig:1d_scale_density} (a) (solid). 
Finally, the linearization of the covariance results in artifacts away from the point of linearization (the center of the Gaussian), as has been shown by several prior works~\cite{jiang24COSV,Zhang2024RaDeGSRD,Dai2024GaussianSurfels,hahlbohm2024htgs,Yu2024GOF,huang20242d}.
Since our method directly integrates Gaussians in ray-space, it doesn't have this issue and the artifacts it causes.

The overall effect is that our method can better fit opaque objects. 
We demonstrate this in \cref{fig:piecewise_constant} where we have fit 3D Gaussians to a single image (\cref{fig:piecewise_constant}(a)). 
We observe that our method (\cref{fig:piecewise_constant}(b)) better approximates sharp edges and constant regions versus 3DGS (\cref{fig:piecewise_constant}(c)), which blurs edges and shows bleeding artifacts. We also validate this observation in a synthetic multiview setting in Supp.C. 

\section{Implementation details}
Our method is implemented in the 3DGS framework using their SLANG.D rasterizer~\cite{bangaru2023slangd}.
We found that this configuration produces better results for 3DGS (and is slightly slower) than the implementation used in their paper; for fair comparison, we use this configuration for both 3DGS and our method.
In the rasterizer, we simply swap out 3DGS's $\alpha$ computation with ours.
\paragraph{Reparameterizing density:} Our method is able to make the density $\kappa$ high to fit opaque shapes well. However, opaque shapes are close to piecewise-constant, resulting in very small gradients, which can hamper optimization. To handle this, similar to concurrent work~\cite{mai2024ever}, we reparameterize the density  
\begin{align}
    \kappa = -\log(1- 0.99\theta) \frac{1}{3}\bigg(\frac{1}{s_x} + \frac{1}{s_y} + \frac{1}{s_z}\bigg),
    \label{eq:density_reparam}
\end{align}
where $\theta\in[0,1]$.
The reparameterization promotes high density for small Gaussians and discourages it for larger ones, which we have found to benefit view synthesis quality.

\paragraph{Heuristics and hyperparameters:} We make a few small changes to 3DGS's adaptive densification process for our method: a) we halve density $\kappa$ after splitting and cloning; b) we also split points with high density $\kappa$; c) we prune points with low density (based on $\theta$ i.e. density before reparameterization); d) we disable opacity reset since we found it didn't help;
We list other hyperparameters and implementation details in Supp.D. 

\section{Applications}
We developed our method primarily for view synthesis. However, since we analytically compute transmittance, our proposed rendering model also works out-of-the-box (with no modifications) for 3DGS-based tomography pipelines, where we match or exceed the state-of-the-art in performance, with fewer points.
Our method is compatible with other extensions of 3DGS, for both view synthesis ~\cite{kheradmand20243d} and downstream applications like sonar reconstruction ~\cite{qu2024z}; we leave these for future work.

\begin{table*}[t]
  \setlength{\tabcolsep}{3pt}
    \centering
    \caption{
          \textbf{Quantitative Comparison.} We compare our approach against prior work on the MipNerf-360, Tanks and Temples ~\cite{knapitsch2017tanks} and DeepBlending~\cite{hedman2018deep} datasets. Metrics for 3DGS-Slang and our method are obtained from our experiments using an Nvidia 3090-Ti GPU. We trained and evaluated ZipNeRF using an Nvidia A-40 GPU. We obtained the numbers directly from the paper authors for concurrent work EVER. All the other baseline numbers are taken from their respective papers. The results are colored as \boxcolorf{best}, \boxcolors{second-best}, and \boxcolort{third-best} among point-based methods (see~\cref{subsec:res_view_synthesis}). We use the same train-test split as ~\citet{kerbl20233d}.}
        \scalebox{0.85}{ 
      \begin{tabular}{l|cccc|cccc|cccc|}
    Dataset & \multicolumn{4}{c|}{Mip-NeRF360} & \multicolumn{4}{c|}{Tanks\&Temples} &  \multicolumn{4}{c|}{DeepBlending} \\ \hline
    Method|Metric 
    & PSNR$\uparrow$ & SSIM$\uparrow$ & LPIPS$^\downarrow$ & FPS 
    & PSNR$\uparrow$ & SSIM$\uparrow$ & LPIPS$^\downarrow$ & FPS 
    & PSNR$\uparrow$ & SSIM$\uparrow$ & LPIPS$^\downarrow$ & FPS \\ \hline 
    Plenoxels ~\cite{yu2021plenoctrees}& 23.08 & 0.626 & 0.463 & 6.79 
    & 21.08 & 0.719 & 0.379 & 13.0 
    & 23.06 & 0.795 & 0.510 & 11.2 \\
    INGP-Big ~\cite{muller2022instant}& 25.59 & 0.699 & 0.331 & 9.43 
    & 21.92 & 0.745 & 0.305 & 14.4 
    & 24.96 & 0.817 & 0.390 & 2.79 \\
    M-NeRF360 ~\cite{barron2022mipnerf360}& 27.69 & 0.792 & 0.237 & 0.06 
    & 22.22 & 0.759 & 0.257 & 0.14 
    & 29.40 & 0.901 & 0.245 & 0.09 \\ \hline
    Zip-NeRF~\cite{barron2023zip}& 29.16 & 0.830 & 0.179 & 0.03 & 28.45 & 0.921 & 0.189 & 0.09 & 30.54 & 0.929 & 0.201 & 0.045 \\ 
    \hline
    Don't Splat~\cite{condor2024dontsplatgaussiansvolumetric} & 22.07 & 0.536 & - & 28-39 
    & \cellcolor{colorf}23.93 & \cellcolor{colort}0.850 & - & 28-39 
    & - & - & - & - \\
    EVER~\cite{mai2024ever} & \cellcolor{colors}27.51 & \cellcolor{colorf} 0.825 & \cellcolor{colorf}0.194 & 36 
    & 23.60 & \cellcolor{colorf} 0.870 & \cellcolor{colorf} 0.160 &  37
    & 29.58 & \cellcolor{colorf}0.908 & 0.308 & 44 \\ \hline 
    3DGS-Slang ~\cite{kerbl20233d} & \cellcolor{colorf}27.52 & \cellcolor{colors}0.813 & \cellcolor{colort}0.222 & 159 
    & \cellcolor{colort}23.64 & \cellcolor{colort}0.850 & \cellcolor{colort}0.176 & 225  
    & \cellcolor{colorf}29.78 & \cellcolor{colors}0.906 & \cellcolor{colors} 0.248 & 163 \\
    GES ~\cite{hamdi2024ges}& 26.91 & \cellcolor{colort}0.794 & 0.250 & 186 
    & 23.35 & 0.836 & 0.198 & 210 
    & \cellcolor{colort}29.68 & \cellcolor{colort} 0.901 & \cellcolor{colort}0.252 & 160 \\
    \textbf{Ours} & \cellcolor{colort}27.30 & \cellcolor{colors}0.813 & \cellcolor{colors}0.209 & 136 
    & \cellcolor{colors}23.74 & \cellcolor{colors}0.854 & \cellcolor{colors}0.167 & 185 
    & \cellcolor{colors}29.72 & \cellcolor{colorf}0.908 & \cellcolor{colorf}0.247 & 172 \\
\end{tabular}}
      \label{tab:comparisons} 
  \end{table*}

\subsection{View synthesis}
\label{subsec:res_view_synthesis}
We evaluate our method on MipNeRF-360, DeepBlending, and Tanks\&Temples Datasets. For DeepBlending and Tanks\&Temples we use the same scenes as 3DGS. We compare with two kinds of baselines.
NeRF and voxel-based methods ~\cite{fridovich2022plenoxels, muller2022instant, barron2022mip, barron2023zip}, and point-based methods (rows below Zip-NeRF in \cref{tab:comparisons}) that include ray-tracing ~\cite{mai2024ever, condor2024dontsplatgaussiansvolumetric}, and rasterization-based 3DGS follow-ups that change the alpha computation similar to what we do ~\cite{hamdi2024ges}.
Many works improve 3DGS along various axes ~\cite{kheradmand20243d, kim2024color} which are orthogonal to our contribution. 
We only show comparisons with methods that directly change the alpha computation and are rasterization-based.

\paragraph{Reconstruction quality:}
We consistently match or outperform rasterization-based methods such as 3DGS  and GES on SSIM and LPIPS.
Since SSIM and LPIPS are more sensitive to edge quality and perceptual similarity, our method performs well on these metrics compared to PSNR, where the ability of 3DGS to render more diffuse textures skews results in its favor.
Our method allows 3D Gaussian primitives to represent opaque textures better than 3DGS, leading to comparable or better quality.
While GES produces fewer primitives, ours has better reconstruction quality and is compatible with the primitives proposed in their work. 
For point-based ray-tracing methods, we have better reconstruction performance than ~\citet{condor2024dontsplatgaussiansvolumetric}, which is closest to our work algorithmically.
EVER ~\cite{mai2024ever} has better reconstruction quality,
but since it uses ray tracing, it has a much lower FPS and higher training times than ours. ZipNeRF~\cite{barron2023zip}
achieves state-of-the-art performance but is prohibitively slow to train and render. We report per-scene metrics from our method on all three datasets in Supp.E.
\paragraph{Runtimes and storage:}
Since our method is rasterization-based, we are faster than all the ray tracing-based methods for rendering images (and similar to 3DGS / GES).
While GES has the lowest storage requirements and highest FPS (though its reconstruction quality is worse than ours).
Our training times are much lower than point-based ray-tracing and NeRF-based methods, but slightly slower than 3DGS, see Supp.F.
The training slowdown is due to more complex computations in the forward and backward pass, and extra computations in the vertex shading phase for our method compared to 3DGS; more details are in Supp.F. We also include details on our ZipNeRF training setup in Supp.F. We show that our method outperforms 3DGS across a wide range of memory budgets in Supp.G. We ablate our method and present the results in Supp.H.
\begin{figure*}[ht!]
    \centering
    \includegraphics[width=0.73\linewidth]
    {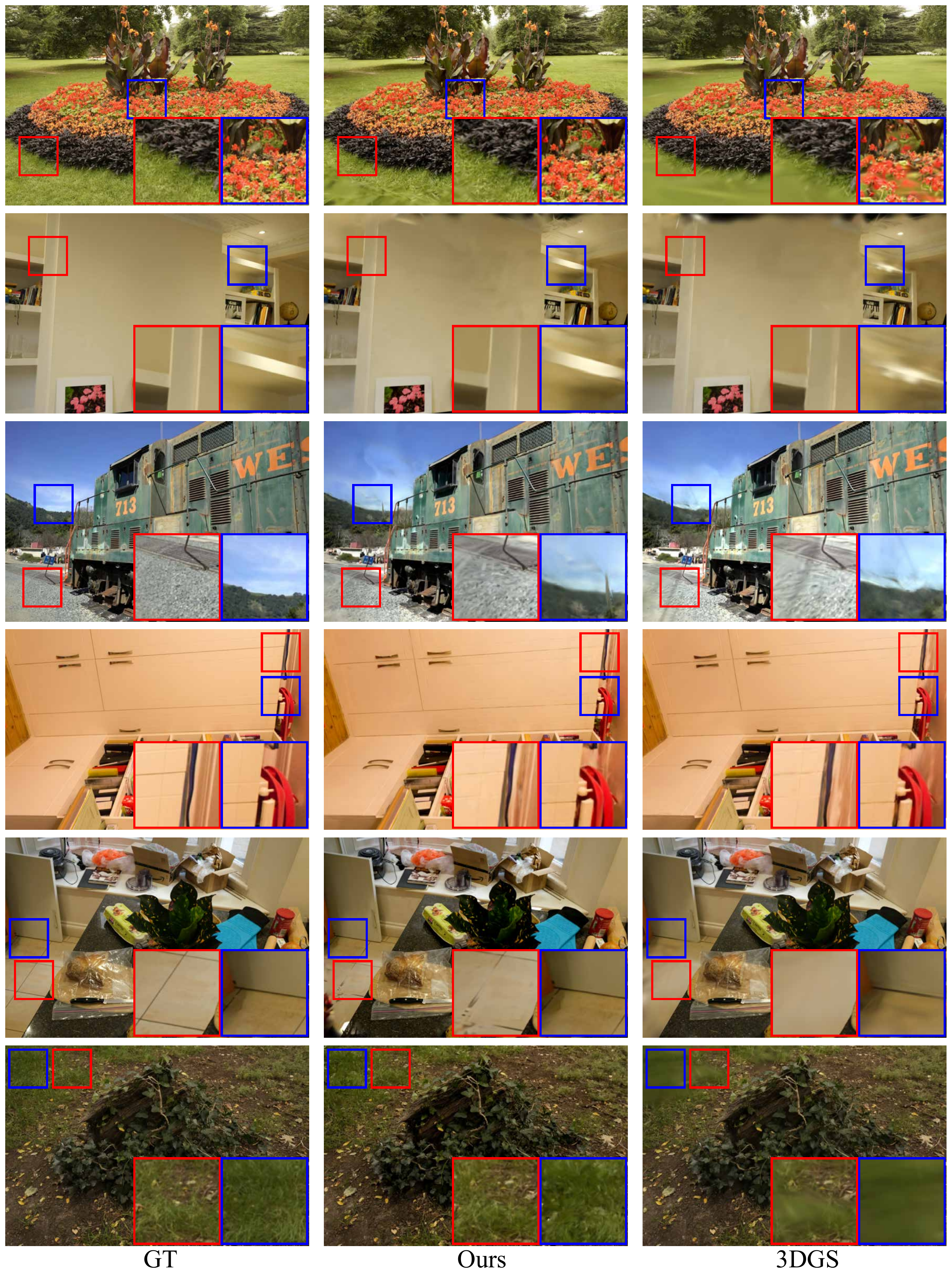}
    \caption{\textbf{Qualitative Results on View Synthesis} We qualitatively show the strengths of our approach on several scenes from real datasets.
    Each row from top to bottom are for the scenes
    \textsc{flowers},
    \textsc{room},
    \textsc{trains},\textsc{playroom},
    \textsc{counter} and \textsc{stump}.  Our method has less blurry regions compared to 3DGS in \textsc{flowers}.
    Our ability to represent opaque regions accurately shows up prominently in \textsc{room}.
    Our method is able to better capture fine details in the tiles in \textsc{playroom}.
    In \textsc{counter} 3DGS misses details on the floor that our method is able to capture.
    Finally, in \textsc{stump}, we are able to capture sharper details in the grass that 3DGS misses.
    }
    \label{fig:viewsynth_results}
\end{figure*}

\subsection{Tomography}
\label{subsec:res_tomography}
We briefly review the goal and setup of computed tomography, see~\cite{zha2024r} for an in-depth review.
The goal is to recover the underlying 3D density field $\sigma(\boldsymbol{x})$ of a scene from 2D images.
The (logarithmic) intensity of a pixel, governed by Beer-Lamberts law
is given by
\begin{align}
    I(p) = \log(I_0) - \log(I') = \int_{t_n}^{t_f} \sigma(\boldsymbol{r}(t))dt.
\end{align}
where $I_0$ is the source intensity and $I'(\boldsymbol{r})$ is the detected intensity.

Using 3DGS out-of-the-box for tomography reconstruction produces poor results.
This is because the splatting process isn't physically accurate, as shown by $R^2$-Gaussian~\citet{zha2024r}.
They identify that unnormalized 3D Gaussians (as used by 3DGS), when splatted onto the screen, are view inconsistent.
To fix this, they devise a specialized rendering model for tomography by normalizing the 3D Gaussians, which achieves state-of-the-art reconstruction results.
Tomography reconstruction is slightly simpler than view synthesis since the forward rendering model does not depend on the ordering of the Gaussians, and there is no emission or color involved, only the transmittance along a ray is measured.
While~\citet{zha2024r} remove the integration bias from the alpha computation process by normalizing Gaussians, their approach still incurs the affine approximation in 3DGS.

\begin{figure}
    \centering
    \includegraphics[width=\linewidth]{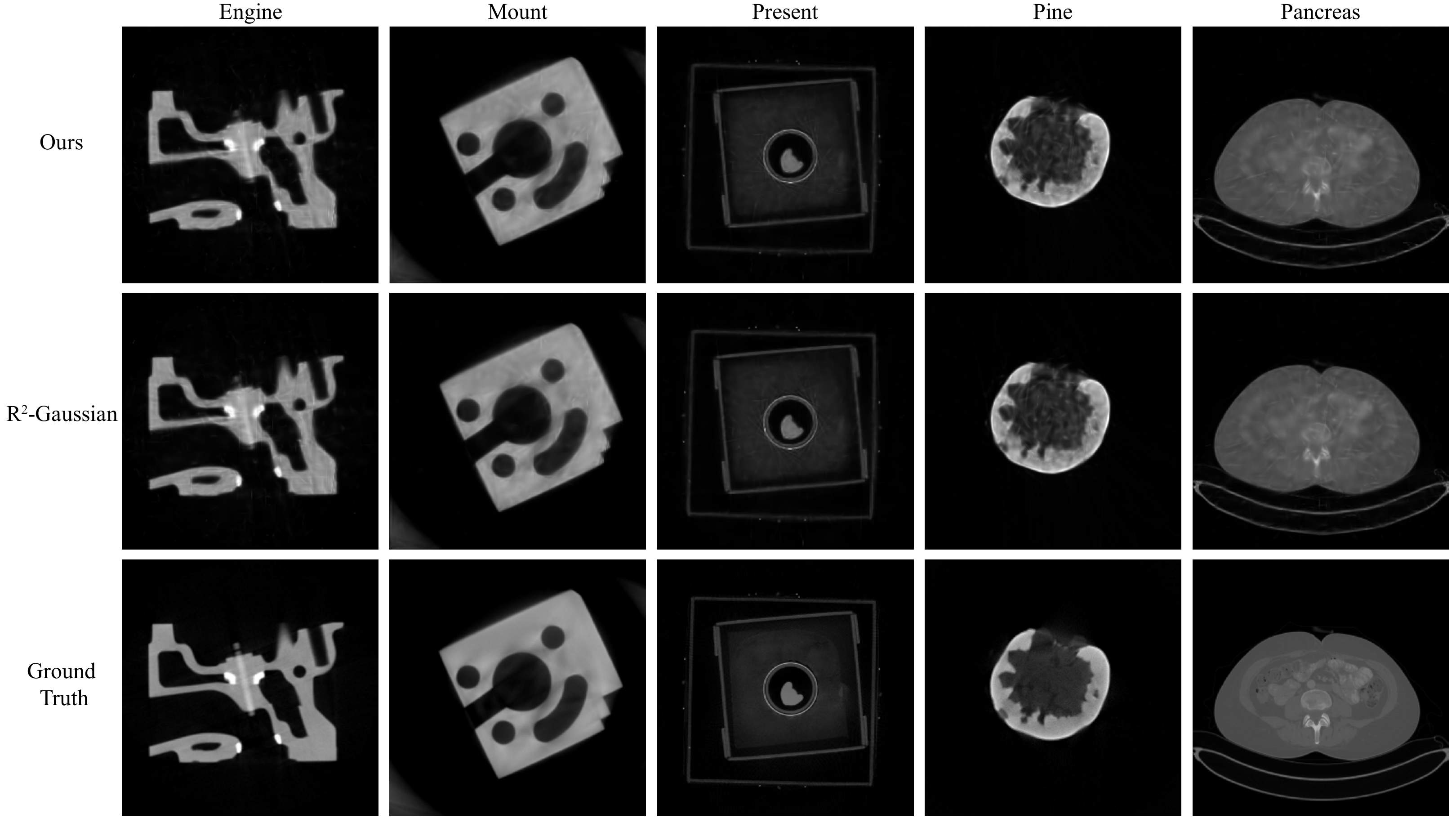}
    \caption{\textbf{Tomography visual results:} For our method (top row) and the splatting-based SOTA $R^{2}$-Gaussian (\citet{zha2024r}), we show slices of the reconstructed 3D density along the axial plane, perpendicular to the frontal views used during training. Each column corresponds to one of the distinct scenes from the datasets in \cite{zha2024r}. Both methods produce visually similar reconstructions, with an average 2D PSNR of 29.94 (ours) and 30.06 (3DGS) for the displayed images.}
    \label{fig:tomo_figure}
\end{figure}

Our method requires no changes to the alpha computation process and still works out of the box for tomography.
We implement our method for tomography by integrating our rasterizer in~\citeauthor{zha2024r}'s codebase and evaluate on their real and synthetic datasets. Since our method provides a more accurate forward rendering model for tomography, we produce fewer points compared to~\citeauthor{zha2024r}, while still matching them in performance on 3D PSNR and 3D SSIM on both real and synthetic datasets.  We show results on their most challenging sparse-view (25 views) reconstruction in~\cref{tab:tomography_results} and a subset of visual results in~\cref{fig:tomo_figure}.
\begin{table}
\centering
\setlength{\tabcolsep}{3pt} 
\renewcommand{\arraystretch}{1} 
\scalebox{0.68}{%
\begin{tabular}{@{}ccccccccccc@{}}
\toprule
& \multicolumn{4}{c}{Synthetic Dataset} & \multicolumn{4}{c}{Real Dataset} \\ \cmidrule(lr){2-5} \cmidrule(lr){6-9}
\multirow{-2}{*}{Methods} & PSNR$\uparrow$ & SSIM$\uparrow$ & Time$\downarrow$ & Pts & PSNR$\uparrow$ & SSIM$\uparrow$ & Time$\downarrow$ & Pts \\ \midrule

FDK~\cite{feldkamp1984practical} & 22.99 & 0.317 & - & - & 23.30 & 0.335 & - & - \\
SART~\cite{andersen1984simultaneous} & 31.14 & 0.825 & \cellcolor{colors}1m47s & - & 31.52 & 0.790 & \cellcolor{colors}1m47s & - \\
ASD-POCS~\cite{sidky2008image}& 30.48 & 0.847 & \cellcolor{colorf}56s & - & 31.32 & 0.810 & \cellcolor{colorf}56s & - \\
IntraTomo~\cite{zang2021intratomo} & \cellcolor{colort}34.68 & \cellcolor{colort}0.914 & 2h19m & - & \cellcolor{colorf}35.85 & \cellcolor{colorf}0.835 & 2h18m & - \\
NAF~\cite{zha2022naf}& 33.91 & 0.893 & 31m1s & - & 32.92 & 0.772 & 51m24s & - \\
SAX-NeRF~\cite{cai2023structure} & 34.33 & 0.905 & 13h3m & - & 33.49 & 0.793 & 13h25m & - \\
$R^{2}$-Gaussian ~\cite{zha2024r} & \cellcolor{colors}34.97 & \cellcolor{colors}0.921 & 18m55s & 66.4K & \cellcolor{colors}35.01 & \cellcolor{colorf}0.835 & 15m57s & 71K \\
\textbf{Ours} & \cellcolor{colorf}35.08 & \cellcolor{colorf}0.922 & \cellcolor{colort}17m39s & 58.2K & \cellcolor{colort}34.98 & \cellcolor{colorf}0.835 & 30m23s & 67.4K \\ \bottomrule

\end{tabular}%
}
\caption{\textbf{Sparse-view (25 views) tomography.} We report 3D-PSNR and 3D-SSIM computed on the estimated density for synthetic and real data. The metrics for other baselines are taken from ~\citet{zha2024r}.
For both ~\citeauthor{zha2024r} and ours, we report results at 30k iterations from our experiments.
}
\label{tab:tomography_results}
\end{table}
\section{Limitations}
\begin{figure}[t]
    \centering\includegraphics[width=\linewidth]{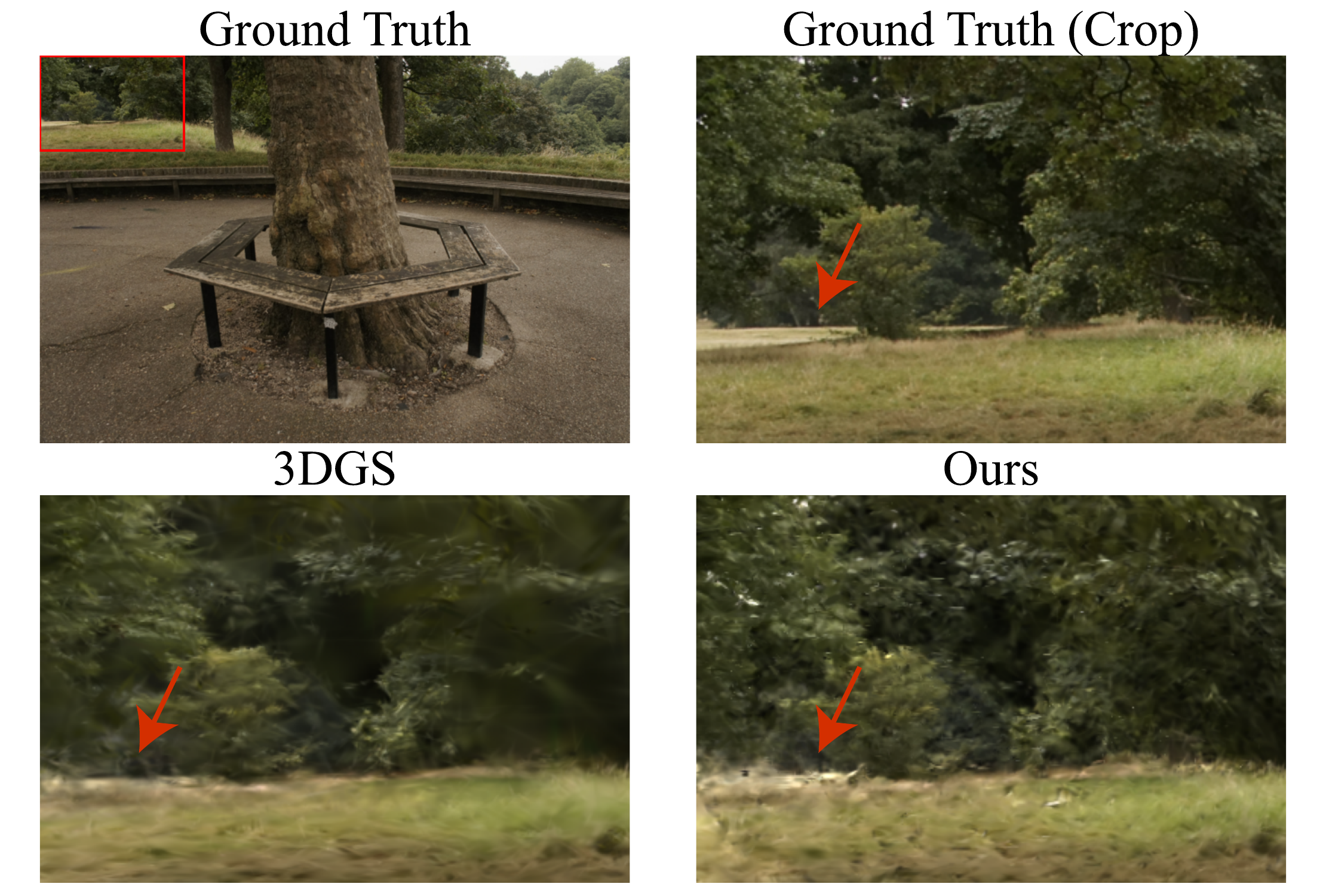}
    \caption{
    \textbf{Opaque artifacts}
    Our method is more sensitive than 3DGS to poor camera calibration and distorted images (Treehill from MipNerf-360).
    We produce opaque artifacts whereas 3DGS prefers a blurry solution with lower error (see red arrows).
    }
    \label{fig:camera_calibration_poor}
\end{figure}

\paragraph{Sorting and overlap:}
Since our method, like 3DGS, is rasterization-based, it inherits some of its issues.
Per-tile (instead of per-pixel) primitive sorting and no overlap handling introduces popping artifacts.
Several recent works ~\cite{radl2024stopthepop,hou2024sort, maule2013hybrid} ameliorate this issue for 3DGS (their methods are readily applicable to us), but they cannot fully eliminate it like ray-tracing-based methods can~\cite{mai2024ever}.
\vspace{-5mm}
\paragraph{Sensitivity to poor calibration:}
In cases of large errors in camera calibration or distortion, our method produces opaque artifacts while 3DGS degrades more gracefully by producing blurry reconstructions, as shown in~\cref{fig:camera_calibration_poor}. Refer to supp.I for a thorough analysis of the failure cases in the MipNerf-360 dataset, which exemplifies this effect.

\section{Future work}
\paragraph{Compact primitives:}
Our method applies to all primitives for which alpha in~\cref{eq:trans} can be evaluated analytically.
Of particular interest are primitives with compact support, like the Epanechnikov kernel, which concurrent work~\cite{condor2024dontsplatgaussiansvolumetric} shows has analytic alpha values, since its compactness minimizes errors due to overlap.
\vspace{-5mm}
\paragraph{Gaussian tile size:}
To compute overlap between primitives and tiles, we use the same heuristics as 3DGS, though this should ideally be adjusted to account for our method's different rendering model.

\section{Conclusion}
We introduce a more physically-accurate version of 3DGS by computing alpha values via analytical 3D Gaussian integration instead of splatting.
Our approach integrates the strengths of physically-based ray tracing with the efficiency of rasterization.
It enables higher quality reconstruction than 3DGS (as measured by SSIM and LPIPS).
We also show that since our method is a plug-and-play replacement for alpha computation in 3DGS, it can be directly used for other 3DGS-based applications such as tomography.

\vspace{-5mm}\paragraph{Acknowledgements:} This work was supported in part by NSF grants 2341952
and 2105806. We also acknowledge ONR grant N0001423-1-2526, gifts from Adobe, Google, Qualcomm and
Rembrand, support from the Ronald L. Graham Chair,
and the UC San Diego Center for Visual Computing. We also thank Alexander Mai, Kaiwen Jiang, and Zachary Novack for helpful discussions and perspectives.
{
    \small
    \bibliographystyle{ieeenat_fullname}
    \bibliography{main}
}

\clearpage
\twocolumn[\begin{center}
    {\LARGE \textbf{Appendix}}
\end{center}]
\appendix


The appendix material is organized as follows. In section \cref{sec:A}, we show a qualitative comparison between our volumetric rasterizer and ray marching to visualize the impact of the sorting and non-overlapping assumptions we make in our method (see Sec. 4.1 in the main paper). 
We derive an expression for the 1D Gaussian distribution along the camera ray in \cref{sec:B} (eq.15 in the main paper). In \cref{sec:C} we validate that our method outperforms 3DGS at approximating sharp edges and constant texture regions in a multiview setting on the nerf-synthetic dataset (Sec. 4 in the main text).
We describe the hyperparameters we use and more details about our modifications to 3DGS's densification strategy in \cref{sec:D} (Sec. 5 of the main paper), and report the reconstruction quality metrics for all the scenes individually in \cref{sec:E}. We report our average training times in \cref{sec:F} (Sec 6.1 of main paper).
In \cref{sec:G} we compare the reconstruction quality (SSIM and LPIPS) of 3DGS and our method across different memory budgets, referenced in Sec. 6.1 of the main paper. 
We present the results of an ablation study on our method in \cref{sec:H}.
We analyze the failure cases from our method compared to 3DGS on the MipNeRF-360 dataset (see Sec. 7 in the main text) and present a quantitative analysis in \cref{sec:I}.

\section{Comparing volumetric rasterizer with ray marching}
\label{sec:A}

\begin{figure*}
    \centering
    \includegraphics[width=\linewidth]{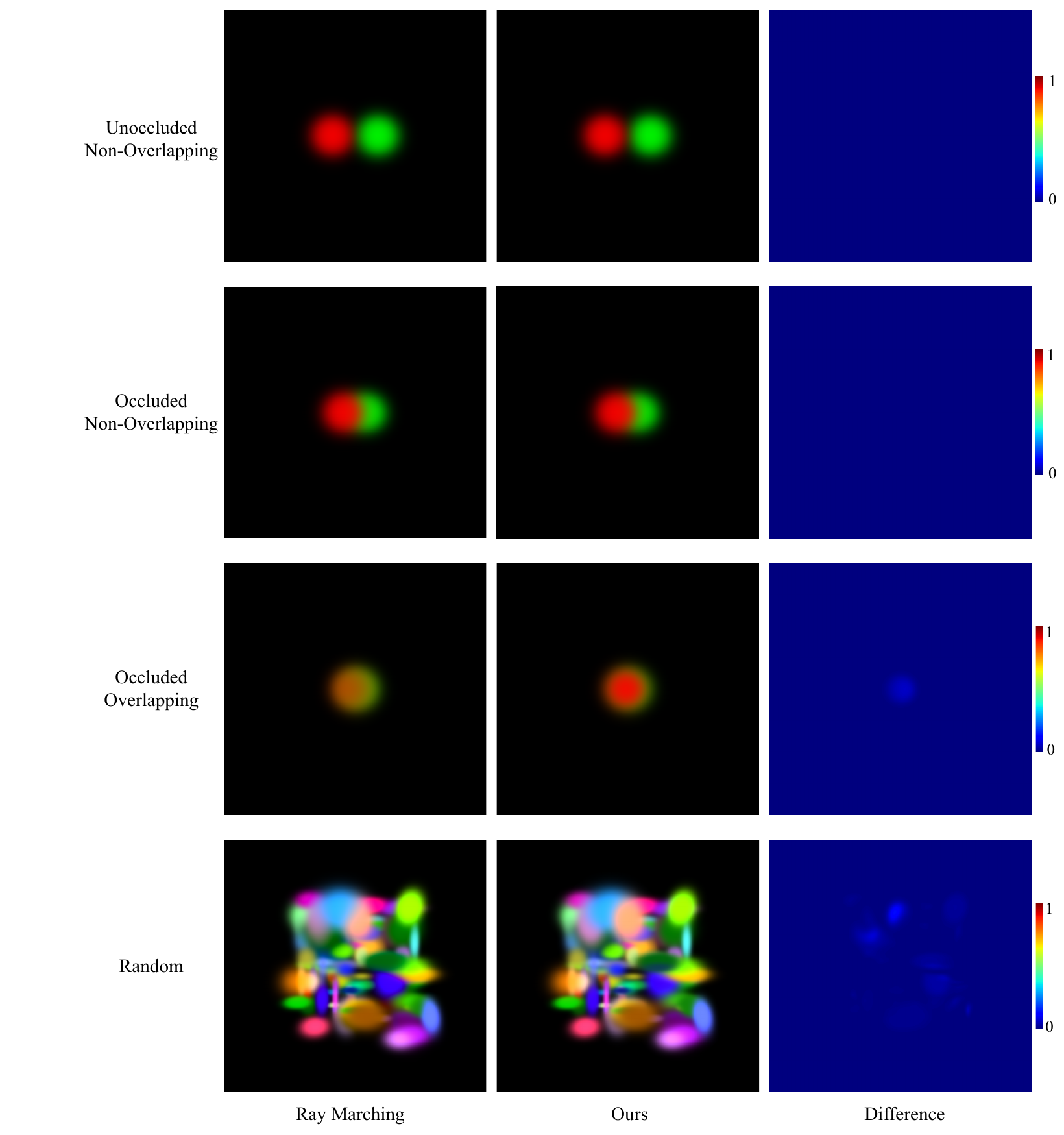}
    \caption{\textbf{Comparison with ray marching} We compare our method with ray marching as a reference for some representative configurations of 3D Gaussians. 
    Since our alpha computation is exact for un-occluded primitives, we match the ray marching result with very high fidelity for non-overlapping Gaussians, in both unoccluded (\textit{Row 1}) and occluded (\textit{Row 2}) cases. Note that in row 2, the primitives occlude each other but are placed far apart along the camera axis ensuring no overlap between them. In (\textit{Row 3}), we place the two primitives very close to each other with a high degree of overlap, resulting in a mismatch between ray marching and our method. To evaluate the effect of the non-overlapping assumption on a more complex setting, we construct 100 Gaussians with uniformly sampled positions and scales in (\textit{Row 4}). Since the primitives are typically smaller than the scene's spatial extent, any overlapping mismatches remain localized. Our method matches the reference accurately in most regions of the image, with minimal differences in areas with overlapping primitives. 
    }
    \label{fig:ray-march-comp}
\end{figure*}
Our method enables volumetrically consistent rasterization by analytically computing the volume rendering integral for each primitive.
Our approach is accurate under the assumption of correctly sorted and non-overlapping primitives (Sec. 3.2 main paper).
In this section, we qualitatively show the effect of this approximation, i.e. to what extent and in which situations it deviates from a reference solution to the volume rendering equation \textit{without approximations}. 
Consequences of overlapping primitives have no relation to view dependence of color, so we assume view-independent colors in this experiment to simplify the ray-marching process. 

We discretize the volume rendering equation using the quadrature rule, similar to eq. 3 in \citet{mildenhall2020nerf} and evaluate it via ray-marching. Given a radiance field parameterized by 3D Gaussians $G_i$ with view independent colors $c_i$, the density $\sigma(\boldsymbol{x})$ and color $c(\boldsymbol{x})$ at the 3D point $\boldsymbol{x}$ are given as:
\begin{align}
    \sigma(\boldsymbol{x}) &= \sum^{N}_{i=1}G_{i}(\boldsymbol{x)} , \;\;\;
    c(\boldsymbol{x}) = \frac{1}{\sigma(\boldsymbol{x})}\sum_{i=1}^{N}c_{i}G_{i}(\boldsymbol{x}).
\end{align}
The color and density above are used to evaluate the volume rendering integral at a pixel by ray-marching along the corresponding ray.
 We compare our method to the ray-marching reference for overlapping and non-overlapping primitives in \cref{fig:ray-march-comp}.
 Our method assumes non-overlapping primitives while computing the volume rendering integral (eq.10, Sec. 4.1 in the main paper), closely matching the ray-marching reference in fidelity in \cref{fig:ray-march-comp} (row 1 and row 2).
 Since our alpha computation assumes non-overlapping primitives, we observe a mismatch between our method and the ray-marching reference when primitives overlap (row 3, \cref{fig:ray-march-comp} ).
 But in practice, the primitives are quite small, and instances of very high overlap are relatively few.
 In row 4 of \cref{fig:ray-march-comp}, we render 100 Gaussian primitives with random means and covariances and observe that our method matches the ray-marching reference closely in all regions apart from those with overlapping primitives, which have a minor impact on the final rendered image.


\section{1D Gaussian along camera ray}
\label{sec:B} 

We outline the proof for eq.15 in the main paper, which describes 
the 3D Gaussian density of a primitive as a 1D Gaussian along a camera ray. 
Recall that in the main paper (eq.4, Sec. 3.2), we defined the 3D Gaussians as
\begin{align}
    G_j(\boldsymbol{x}) = \exp\left\{-\frac{1}{2}(\boldsymbol{x} - \boldsymbol{\mu_j})^{T}\boldsymbol{\Sigma_j}^{-1}(\boldsymbol{x}-\boldsymbol{\mu_j})\right\} \label{eq:pdf3d}
\end{align}
The 1D distribution $g_j(t)$ along the ray $\boldsymbol{x} = \boldsymbol{o} + t\boldsymbol{d}$ is given by (eq. 15 main paper): 

\begin{align}
    g_j(t) &= G_j(\gamma_j \boldsymbol{d}) \exp\left\{\frac{-(t - \gamma_j)^{2}}{2\beta_j^{2}}\right\},\label{eq:1d_gauss}
\end{align}
where $G_j(\gamma_j\boldsymbol{d})$ is the maximum value of the Gaussian along the ray, and the 1D Gaussian has mean $\gamma_j$ and variance $\beta_j$, which are defined as 

\begin{align}
    \gamma_j &= \frac{(\boldsymbol{\mu_j} - \boldsymbol{o})^{T} \Sigma_j^{-1} \boldsymbol{d}}{\boldsymbol{d}^{T}\Sigma_j^{-1}\boldsymbol{d}}, \;\;\;\;\;
    \beta_j = \frac{1}{\sqrt{\boldsymbol{d}^T \Sigma_j^{-1} \boldsymbol{d}}}. \label{eq:betagamma}
\end{align}

To derive this result, we substitute $\boldsymbol{x} = \boldsymbol{o} + t\boldsymbol{d}$ in \cref{eq:1d_gauss}:
\begin{align}
    g_j(t) &= G_j(\boldsymbol{o} + t\boldsymbol{d}) = \exp\left\{-\frac{1}{2}\Delta\right\}, 
\end{align}
where the argument of the exponent,
    \begin{align}
           \Delta &= \bigg(t\boldsymbol{d} - (\boldsymbol{\mu_j} - \boldsymbol{o})\bigg)^{T}\boldsymbol{\Sigma_j}^{-1}\bigg(t\boldsymbol{d}-(\boldsymbol{\mu_j - o})\bigg).
           \end{align}
The expression for $\Delta$ can be further expanded as 
           \begin{align}
           \Delta &= t^{2} \boldsymbol{d}^{T}\boldsymbol{\Sigma_j}^{-1}\boldsymbol{d} - 2t\big(\boldsymbol{\mu_j -o}\big)^{T}\boldsymbol{\Sigma_j}^{-1}\boldsymbol{d} \\ &+ \big(\boldsymbol{\mu_j -o}\big)^{T}\boldsymbol{\Sigma_j}^{-1}\big(\boldsymbol{\mu_j - o}\big)
           \end{align}
           
           \begin{align}
           \textrm{Setting}~K =\frac{\big(\boldsymbol{\mu_j -o}\big)^{T}\boldsymbol{\Sigma_j}^{-1}\big(\boldsymbol{\mu_j - o}\big)}{\boldsymbol{d}^{T}\boldsymbol{\Sigma_j}^{-1}\boldsymbol{d}}  - \gamma^{2}_j,
           \end{align}
           simplifies the expression for $\Delta$ as follows -
           \begin{align}
           \Delta &= \frac{1}{\beta^{2}_j} \bigg(t^2  - 2t\gamma_j +\gamma^{2}_j + K \bigg) \\
           \implies \Delta &= \frac{1}{\beta^{2}_j} (t - \gamma_j)^2 + K \\
           \implies g_j(t) &= \exp\left\{\frac{-(t-\gamma_j)^2}{2\beta^{2}_j} \right\}\exp\left\{-\frac{K}{2}\right\}. \\
\end{align}

In the above equations, $K$ is independent of $t$. Let $t=t_{\max}$ maximize $g_j(t)$. Above, we showed that $$g_j(t) = G_j(\boldsymbol{o} + t\boldsymbol{d}) = \exp\left\{\frac{-(t-\gamma_j)^2}{2\beta^{2}_j} \right\}\exp\left\{-\frac{K}{2}\right\}.$$ Since $\exp\left\{\frac{-(t-\gamma_j)^2}{2\beta^{2}_j} \right\}$ attains its maximum value of 1 when $t = \gamma_j$, $t_{\max} = \gamma_j$. This can also be verified by setting $\frac{d\Delta}{dt}=0$ and solving for $t$. This gives us 
\begin{align}
g_j(t_{\max}) = G_j(\boldsymbol{o}+t_{\max} \boldsymbol{d}) = \exp\left\{-\frac{K}{2}\right\}.
\end{align}

We denote $G_j(\boldsymbol{o} + t_{\max}\boldsymbol{d})$  as $G_j(\gamma_j\boldsymbol{d})$ in the main paper (eq. 15). This concludes the proof of  \cref{eq:1d_gauss} (eq. 15 in main paper). 

\section{Comparing our method and 3DGS on nerf-synthetic}
\label{sec:C}
\begin{figure}
\includegraphics[width=\linewidth]{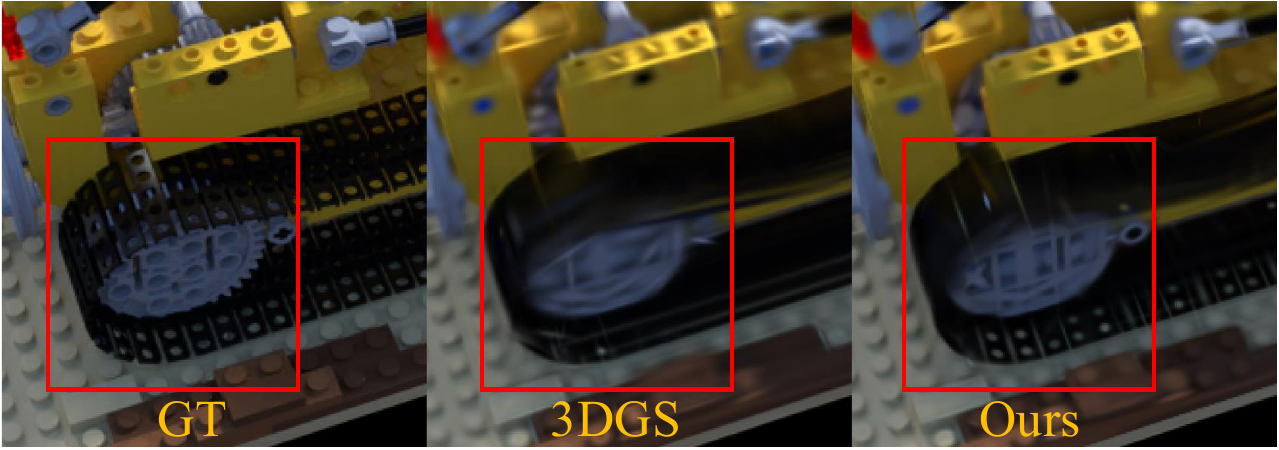}    \caption{\textbf{NerfSynthetic qualitative result.} Our approach renders sharper details for \textit{piecewise opaque regions} (inset) compared to 3DGS (zoom in).}
    \label{fig:legorebuttal}
\end{figure}

Recall that in Fig.2 and Fig.3 in the main text, we fit a single image with a fixed number of primitives and observe that our method is better than 3DGS at approximating sharp edges and constant regions. Here, we validate this observation in a multi-view setting on the nerf-synthetic (NS) dataset. We optimize our method and 3DGS with the \textit{same initialization, and no densification} to isolate the expressiveness of our method from the optimization dynamics.   
For all metrics (PSNR, SSIM, LPIPS) we (29.81, 0.941, 0.074) outperform 3DGS (29.56, 0.936, 0.082) averaged over all 8 scenes in NS. Our method renders piece-wise constant textures (common in nerf-synthetic) better than 3DGS.
See \cref{fig:legorebuttal} for a qualitative comparison. 

\section{Implementation details and hyperparameters}
\label{sec:D}
\paragraph{Implementation details:} Unlike 3DGS which uses 2D screen space positional gradients, we use the world space 3D positional gradients to compute the gradient norm for densification thresholding, similar to \cite{moenne20243d, mai2024ever}.
We use separate positional gradient thresholds for the splitting $(10^{-8})$ and cloning $(5\times10^{-5})$ operations. 
We require much lower gradient threshold values since our method produces more opaque primitives than 3DGS which reduces the positional gradient value.
We densify and prune every 200 iterations. For pruning, we prune based on the value of $\theta$, the parameter used to parameterize the density of the primitives $\kappa$ (see eq. 20 in the main paper). We remove primitives for which $\theta$ is less than a minimum opacity threshold. Similar to \cite{mai2024ever}, we apply softplus activation ($\beta=10$) to the spherical harmonic coefficients.
%
\paragraph{Hyperparameters:} The learning rates for density and color features are set to $0.03$ and $0.003$ respectively. Initial and final position learning rates are set to $0.00016$ and $0.00001$.
In the 3DGS codebase, the scene extent (or camera extent) is defined as the maximum distance between any training camera and the average camera centroid. The scene extent controls the splitting, cloning, and pruning thresholds in the densification process. In our method, we scale the scene extent by a factor of $2, 5$, and $1.5$ for the Mip-NeRF360, Tanks\&Temples, and DeepBlending datasets respectively. We also adjust the (hardcoded) pruning hyperparameter to prune points with large scales to 0.01 from the default value 0.1. 

\section{Per scene metrics}
\label{sec:E}
\begin{table}[t]
\centering
\begin{tabular}{|l|c|c|c|c|}
\hline
\textbf{Scene} & \textbf{PSNR} & \textbf{SSIM} & \textbf{LPIPS} & \textbf{\#Points} \\ \hline
Bicycle       & 25.04         & 0.7594        & 0.2155         & 3.7M             \\ \hline
Bonsai        & 32.17         & 0.9498        & 0.1704         & 1.5M             \\ \hline
Counter       & 28.92         & 0.9206        & 0.1699         & 1.6M             \\ \hline
Garden        & 27.29         & 0.858        & 0.122         & 3.18M            \\ \hline
Flowers       & 21.41         & 0.615         & 0.306          & 3.8M             \\ \hline
Stump         & 26.70         & 0.778         & 0.219          & 5.3M             \\ \hline
Treehill      & 21.94         & 0.627         & 0.329          & 3.6M             \\ \hline
Room          & 32.00         & 0.9358        & 0.1845         & 1.6M             \\ \hline
Kitchen       & 31.51         & 0.935         & 0.113          & 2.1M             \\ \hline
\end{tabular}
\caption{Metrics on MipNeRF360 scenes for our method.}
\label{tab:scene_metrics_mip}
\end{table}

\begin{table}[t]
\centering
\begin{tabular}{|l|c|c|c|c|}
\hline
\textbf{Scene} & \textbf{PSNR} & \textbf{SSIM} & \textbf{LPIPS} & \textbf{\#Points} \\ \hline
Train         & 22.16         & 0.825         & 0.195          & 1.91M            \\ \hline
Truck         & 25.31         & 0.8838        & 0.139          & 1.54M            \\ \hline
\end{tabular}
\caption{Metrics on Tanks\&Temples scenes for our method.}
\label{tab:scene_metrics_tandt}
\end{table}

\begin{table}[t]
\centering
\begin{tabular}{|l|c|c|c|c|}
\hline
\textbf{Scene} & \textbf{PSNR} & \textbf{SSIM} & \textbf{LPIPS} & \textbf{\#Points} \\ \hline
Playroom      & 30.23         & 0.909         & 0.250          & 2.2M             \\ \hline
DrJohnson     & 29.22         & 0.906         & 0.244          & 4.7M             \\ \hline
\end{tabular}
\caption{Metrics on DeepBlending scenes for our method.}
\label{tab:scene_metrics_db}
\end{table}
We report metrics from our method for each scene individually in Mip-NeRF360, Tanks\&Temples, and DeepBlending datasets in \cref{tab:scene_metrics_mip}, \cref{tab:scene_metrics_tandt} and \cref{tab:scene_metrics_db}. Note that in the main paper (Tab. 1), we report metrics averaged over all scenes for the three datasets -Mip-NeRF360, Tanks\&Temples, and DeepBlending. For per-scene metrics on some of the
other baselines, please see \cite{kerbl20233d, hamdi2024ges}.

\section{Average training times}
\label{sec:F}
We use a Nvidia 3090 Ti for training both our method and 3DGS. Our training times are similar to 3DGS, with a slight slowdown that arises from the extra computations needed for our alpha computation compared to 3DGS. 3DGS is able to re-use the splatted 2D Covariance matrix for computing alpha for each pixel. Our method requires computing $t_{\max}$ and other intermediaries for each pixel separately which leads to a slight slowdown. We compute inverses of both the 2D and 3D covariances, as opposed to 3DGS which does the inverse computation only once for the 2D covariance in the vertex shader phase. The vertex shader in our slang.D implementation uses atomic add operations to write the computed inverse covariance matrices to global memory. Our method requires more atomic adds per primitive to store 3D covariance matrices, leading to slower performance compared to 3DGS, which requires fewer atomic adds to store 2D covariance matrices.

For MipNeRF-360 scenes, on average 3DGS and our method generate 2.62M and 2.93M points respectively, and the training times for 30000 iterations are 51.5 min and 61 min respectively. Similarly for Tanks\&Temples, on average 3DGS and our method generate 1.79M and 1.72M points respectively, and the training times for 30000 iterations are 20.5 min and 33.2 min respectively.
\paragraph{ZipNerf training:} We trained and evaluated ZipNeRF~\cite{barron2023zip} on a Nvidia A-40 GPU, using their official code release~\cite{campzipnerf2024}. In the official code release for ZipNeRF, the indoor and outdoor scenes in the Mip-NeRF360 dataset are evaluated at 2x and 4x downsampled resolutions respectively.
We trained and evaluated ZipNeRF at full resolution on the Tanks\&Temples and DeepBlending datasets, but for the MipNeRF-360 dataset, we used 2x downsampled images (for both indoor and outdoor scenes) to fit them in the GPU memory. ZipNeRF took approximately 12-13 hours per scene to train on our GPU. For the Tanks\&Temples dataset, we re-ran COLMAP before training Zip-NeRF as highlighted in issue 7 in the official Zip-NeRF code release~\cite{campzipnerf2024}.
\section{Reconstruction quality for different memory budgets}
\label{sec:G}
\begin{figure}
    \centering
    \includegraphics[width=\linewidth]{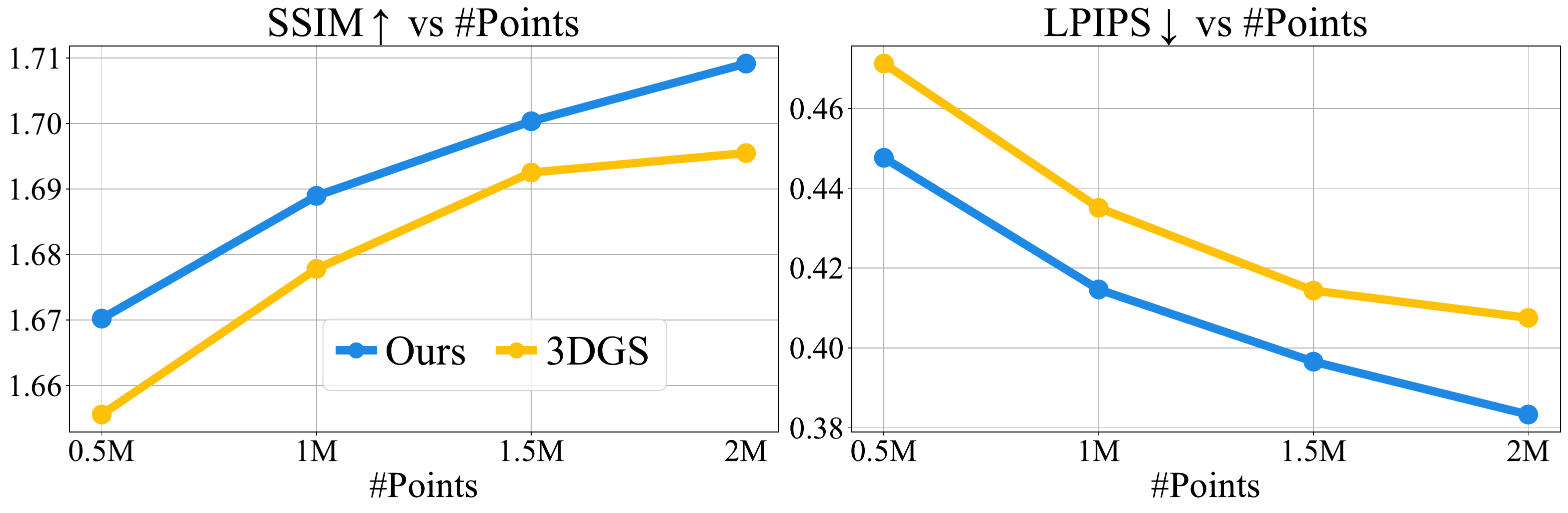}
    \caption{\textbf{Varying number of primitives}: Each data point is computed by averaging the test dataset metrics over the scenes \textsc{kitchen, stump, train and counter}}
    \label{fig:rate_distortion}
\end{figure}

In \cref{fig:rate_distortion} we compare the performance of our method against 3DGS across different memory budgets, i.e. across a different number of maximum primitives that each method is allowed to generate.
We stop densification and pruning once the optimization reaches the maximum number of allowed primitives. For each memory budget, we average the results over 4 scenes, \textsc{kitchen, stump, train, counter}. For the same memory budget, our method consistently outperforms 3DGS on both SSIM and LPIPS. This holds across a range of memory budgets as shown in \cref{fig:rate_distortion}. Recall that our method represents opaque textures better than 3DGS for the same number of primitives. We demonstrated this through the toy examples in Fig. 2 and Fig. 3 in the main paper. Since LPIPS and SSIM measure perceptual similarity and edge quality, this also translates to quantitative improvements as demonstrated in \cref{fig:rate_distortion}. 
\section{Ablation study}
\label{sec:H}
\begin{table}[t]
    \centering
    \resizebox{\columnwidth}{!}{%
    \begin{tabular}{l|r|r|r|r|c}
                   & \multicolumn{1}{l}{PSNR $\uparrow$} & \multicolumn{1}{l}{SSIM $\uparrow$} & \multicolumn{1}{l|}{LPIPS $\downarrow$} &\multicolumn{1}{l}{Points}               \\ \hline
3DGS & \cellcolor{colors}21.38 & \cellcolor{colort}0.588 & 0.360 & 3.4M\\
3DGS + Our HyperParams & 20.66 & 0.571 & \cellcolor{colort} 0.329 & 7.4M \\
Ours w/o Reparam  &  \cellcolor{colort} 21.23 & \cellcolor{colors}0.600 & \cellcolor{colors} 0.309 & 7M \\
Ours w EVER Reparam & 20.04 & 0.519 & 0.396 & 9.4M \\
Ours  & \cellcolor{colorf} 21.41	& \cellcolor{colorf}0.615	& \cellcolor{colorf}0.306	& 3.8M \\ 
\hline
\end{tabular}%
}
    \caption{\textbf{Ablations}. We show ablations on the \textsc{flowers} scene in MipNeRF-360 dataset.}
    \label{tab:ablations}
\end{table}
We conduct an ablation study to assess the impact of hyperparameters and density parameterization on our method, with results shown in \cref{tab:ablations}. We compare our method with 3DGS (row 1), and for fairness, we also run 3DGS with the hyperparameters used by our method. Using 3DGS with our hyper-parameters (row 2) improves LPIPS, but worsens PSNR and SSIM while generating almost 2x more primitives compared to the base 3DGS configuration (row 1). Our method (row 4) fares relatively better on all metrics. 

We also run our method without density reparameterization (row 3); we ensure positive density by applying the softplus activation.
We observe improvement in LPIPS and SSIM compared to 3DGS but at almost double the number of primitives. We also experiment with the parameterization proposed in \citet{mai2024ever} (row 4), which results in poorer SSIM and LPIPS while producing too many points compared to our method (row 5). 

\begin{table*}
\centering
\resizebox{0.6\textwidth}{!}{
\begin{tabular}{|l|c|c|c|c|c|c|}
\hline
Dataset  & Worst Gap  LPIPS & Better / Comparable \% & Failure \% & Total Images \\
\hline
flowers  & -0.022 & 100.0 & 0.000  & 21 \\
room     &  0.005 & 97.4  & 2.632  & 38 \\
bonsai   &  0.004 & 100.0 & 0.000  & 36 \\
bicycle  & -0.008 & 100.0 & 0.000  & 24 \\
treehill &  0.017 & 88.2  & 11.765 & 17 \\
counter  & -0.006 & 100.0 & 0.000  & 29 \\
stump    &  0.004 & 100.0 & 0.000  & 15 \\
garden   &  0.007 & 91.3  & 8.696  & 23 \\
kitchen  &  0.004 & 100.0 & 0.000  & 34 \\
\hline
\end{tabular}}
\resizebox{0.6\textwidth}{!}{
\begin{tabular}{|l|c|c|c|c|c|c|}
\hline
Dataset  & Worst Gap SSIM  & Better / Comparable \% & Failure \% & Total Images \\
\hline
flowers  &  0.005  & 100.0 & 0.000  & 21 \\
room     & -0.004  & 100.0 & 0.000  & 38 \\
bonsai   & -0.006  & 100.0 & 0.000  & 36 \\
bicycle  & -0.002  & 100.0 & 0.000  & 24 \\
treehill & -0.035  & 58.8  & 41.176 & 17 \\
counter  & -0.001  & 100.0 & 0.000  & 29 \\
stump    & -0.013  & 93.3  & 6.667  & 15 \\
garden   & -0.005  & 100.0 & 0.000  & 23 \\
kitchen  & -0.007  & 100.0 & 0.000  & 34 \\
\hline
\end{tabular}}
\resizebox{0.6\textwidth}{!}{
\begin{tabular}{|l|c|c|c|c|c|c|}
\hline
Dataset  & Worst Gap PSNR & Better / Comparable \% & Failure \% & Total Images \\
\hline
flowers  & -1.204  & 95.2  & 4.762  & 21 \\
room     & -1.152  & 97.4  & 2.632  & 38 \\
bonsai   & -2.328  & 94.4  & 5.556  & 36 \\
bicycle  & -0.414  & 100.0 & 0.000  & 24 \\
treehill & -1.174  & 88.2  & 11.765 & 17 \\
counter  & -2.163  & 93.1  & 6.897  & 29 \\
stump    & -0.636  & 100.0 & 0.000  & 15 \\
garden   & -0.536  & 100.0 & 0.000  & 23 \\
kitchen  & -5.764  & 91.2  & 8.824  & 34 \\
\hline
\end{tabular}}
\caption{\textbf{Mip-NeRF-360 dataset quantitative analysis} Worst Gap is the largest negative difference between our result and 3DGS amongst test views for a scene, indicating the frame where our method performed the worst relative to 3DGS. For LPIPS (\textit{top table}), a lower worst gap is better, whereas for SSIM (\textit{middle table}) and PSNR (\textit{bottom table}), a higher worst gap is preferable. As measured by LPIPS (\textit{top table}) and SSIM (\textit{middle table}), our approach is better or comparable (LPIPS within 0.005, SSIM within 0.01) to 3DGS   for most scenes in Mip-NeRF360 except \textsc{treehill}. 
As measured by PSNR (\textit{bottom table}), we observe that our approach suffers the most compared to 3DGS for \textsc{kitchen}, \textsc{counter}, and \textsc{bonsai}. A few anomalies cause a high worst gap, though most views are better or comparable to 3DGS. For all other Mip-NeRF360 dataset scenes, we are comparable (PSNR within 1dB) to 3DGS. }

\label{tab:mipnerf360_results_failure}
\end{table*}




\begin{figure}
    \centering
    \includegraphics[width=\linewidth]{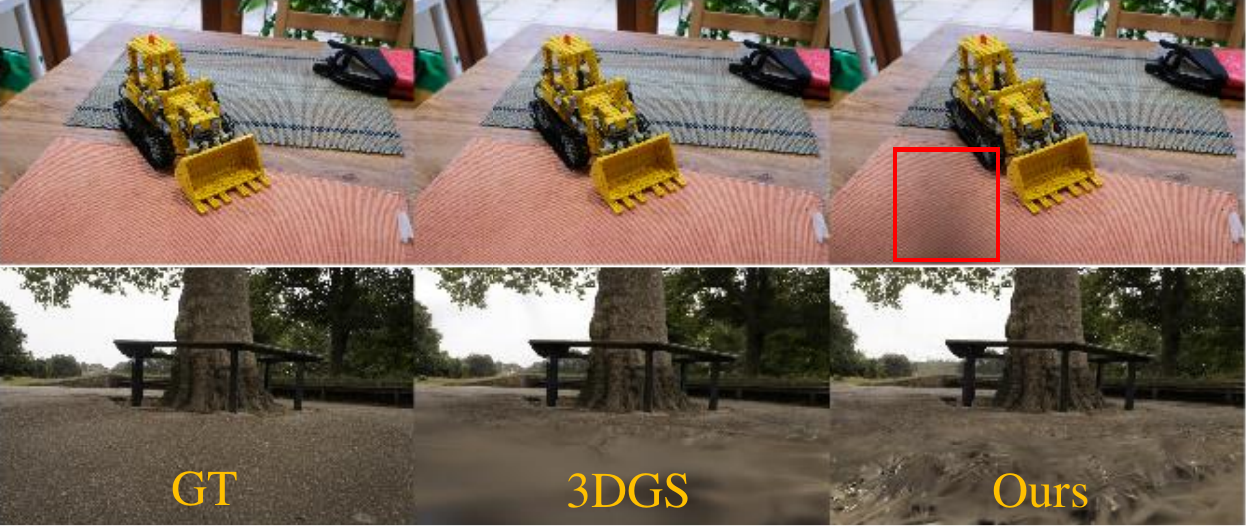}
    \caption{\textbf{MipNerf-360 dataset worst failure cases.} Top row (\textsc{kitchen}): massive floater (inset) causes PSNR drop. Bottom row (\textsc{treehill}): failure due to insufficient densification and possibly inaccurate camera poses. 3DGS prefers a blurry solution, ours has sharp opaque artifacts.}
    \label{fig:mipnerf360-failure}
\end{figure}

\section{MipNeRF-360 dataset failure case analysis}
\label{sec:I}
We analyze failure cases for the 3 metrics. We define a failure case as being worse than 3DGS by a threshold, separately chosen for each metric (1dB PSNR, 0.01 SSIM, 0.005 LPIPS). Averaged on all test images in the dataset, our failure rate is 4.64\%, 3.38\%, and 2.11\% for PSNR, SSIM and LPIPS respectively. \textsc{treehill} shows the highest failure rate among all scenes, with LPIPS and SSIM differences of 0.017 and 0.027, respectively (example in \cref{fig:mipnerf360-failure}). For PSNR, we observe an outlier failure case from \textsc{kitchen}, 5.6 dB worse than 3DGS (\cref{fig:mipnerf360-failure}). However, averaged over all 34 \textsc{kitchen} test images, our method (31.51 dB) performs only slightly worse than 3DGS (31.67 dB) in PSNR. See \cref{tab:mipnerf360_results_failure} for per-scene quantitative analysis on all three metrics for Mip-NeRF360 dataset. 
\newpage

\end{document}